\documentclass[preprint]{vgtc}               % preprint

\ieeedoi{10.1109/VAST47406.2019.8986940}

\ifpdf%                                % if we use pdflatex
  \pdfoutput=1\relax                   % create PDFs from pdfLaTeX
  \usepackage{graphicx}                % allow us to embed graphics files
  \DeclareGraphicsExtensions{.pdf,.png,.jpg,.jpeg} % for pdflatex we expect .pdf, .png, or .jpg files
\else%                                 % else we use pure latex
  \usepackage{graphicx}                % allow us to embed graphics files
  \DeclareGraphicsExtensions{.eps}     % for pure latex we expect eps files
\fi%

\graphicspath{{figures/}{pictures/}{images/}{./}} % where to search for the images

\usepackage{microtype}                 % use micro-typography (slightly more compact, better to read)
\PassOptionsToPackage{warn}{textcomp}  % to address font issues with \textrightarrow
\usepackage{textcomp}                  % use better special symbols
\usepackage{mathptmx}                  % use matching math font
\usepackage{times}                     % we use Times as the main font
\usepackage{cite}                      % needed to automatically sort the references
\usepackage{tabu}                      % only used for the table example
\usepackage{booktabs}                  % only used for the table example
\usepackage{enumerate}
\usepackage{paralist}
\usepackage{multirow}
\usepackage{booktabs}

\DeclareMathAlphabet{\mathpzc}{OT1}{pzc}{m}{it}

\usepackage[leqno]{amsmath}
\usepackage{amssymb}
\newcommand{\leqnomode}{\tagsleft@true}

\usepackage[usenames,dvipsnames,table]{xcolor}
\usepackage{color}

\usepackage{tikz}
\usepackage{wrapfig}
\graphicspath{{figures/}{pictures/}{images/}{./}} % where to search for the images% % % % % % % % % 
\usepackage{lettrine}

\usepackage{calc}

\newlength\myheight
\newlength\mydepth
\settototalheight\myheight{Xygp}
\usepackage{tcolorbox}

\definecolor{hl1}{HTML}{00114D}
\definecolor{repr}{HTML}{0288D1}

\newtcbox{\myboxrepr}{nobeforeafter,colframe=repr,colback=repr!70!white,boxrule=0.5pt,arc=2pt,
  boxsep=0pt,left=2pt,right=2pt,top=2pt,bottom=2pt,tcbox raise base,fontupper=\color{white}}
\definecolor{rel}{HTML}{536DFE}
\newtcbox{\myboxrel}{nobeforeafter,colframe=rel,colback=rel!70!white,boxrule=0.5pt,arc=2pt,
  boxsep=0pt,left=2pt,right=2pt,top=2pt,bottom=2pt,tcbox raise base,fontupper=\color{white}}  
\definecolor{fs}{HTML}{FF5722}
\newtcbox{\myboxfs}{nobeforeafter,colframe=fs,colback=fs!70!white,boxrule=0.5pt,arc=2pt,
  boxsep=0pt,left=2pt,right=2pt,top=2pt,bottom=2pt,tcbox raise base,fontupper=\color{white}}  
\definecolor{ml}{HTML}{E64A19}
\newtcbox{\myboxml}{nobeforeafter,colframe=ml,colback=ml!70!white,boxrule=0.5pt,arc=2pt,
  boxsep=0pt,left=2pt,right=2pt,top=2pt,bottom=2pt,tcbox raise base,fontupper=\color{white}}  
\definecolor{guide}{HTML}{BDBDBD}
\newtcbox{\myboxguidance}{nobeforeafter,colframe=guide,colback=guide!70!white,boxrule=0.5pt,arc=2pt,
  boxsep=0pt,left=2pt,right=2pt,top=2pt,bottom=2pt,tcbox raise base,fontupper=\color{white}}  

\AtBeginDocument{
}

\usepackage{xspace}
\newcommand{\tool}[1]{\textsc{FDive\xspace}}
\newcommand{\ASplus}{$\mathpzc{L^+}$\xspace}
\newcommand{\ASminus}{$\mathpzc{L^-}$\xspace}

\usepackage{tikz}

\newcommand{\tec}[1]{
\begin{tikzpicture}
    \def\value{#1}
    \pgfmathparse{(0.45-(0.5*\value))+0.57}
    \definecolor{tmpcolor}{rgb}{\pgfmathresult,\pgfmathresult,\pgfmathresult}
    \node[fill=tmpcolor] at (0,0) {#1};
\end{tikzpicture}
}

\newcommand{\tecb}[1]{
\begin{tikzpicture}
    \def\value{#1}
    \pgfmathparse{(0.45-(0.5*\value))+0.57}
    \definecolor{tmpcolor}{rgb}{\pgfmathresult,\pgfmathresult,\pgfmathresult}
    \node[fill=tmpcolor] at (0,0) {\textbf{#1}};
\end{tikzpicture}
}

\newcommand{\tecw}[1]{
\begin{tikzpicture}
    \node[black] at (0,0) {#1};
\end{tikzpicture}
}

\onlineid{1125}

\vgtccategory{Research}

\vgtcinsertpkg

\title{FDive: Learning Relevance Models using Pattern-based \\ Similarity Measures}

\author{Frederik L. Dennig$^1$, Tom Polk$^1$, Zudi Lin$^2$, Tobias Schreck$^3$, Hanspeter Pfister$^2$, and Michael Behrisch$^2$ \\ \scriptsize \centering \hfill $^1$University of Konstanz, Germany\thanks{e-mail: \{frederik.dennig, thomas.polk\}@uni-konstanz.de} \hfill $^2$Harvard University, USA\thanks{e-mail: \{linzudi, pfister, behrisch\}@seas.harvard.edu.} \hfill $^3$Graz University of Technology, Austria\thanks{e-mail: tobias.schreck@cgv.tugraz.at} \hfill} % \textit{Member, IEEE}

\teaser{
  \includegraphics[width=\linewidth]{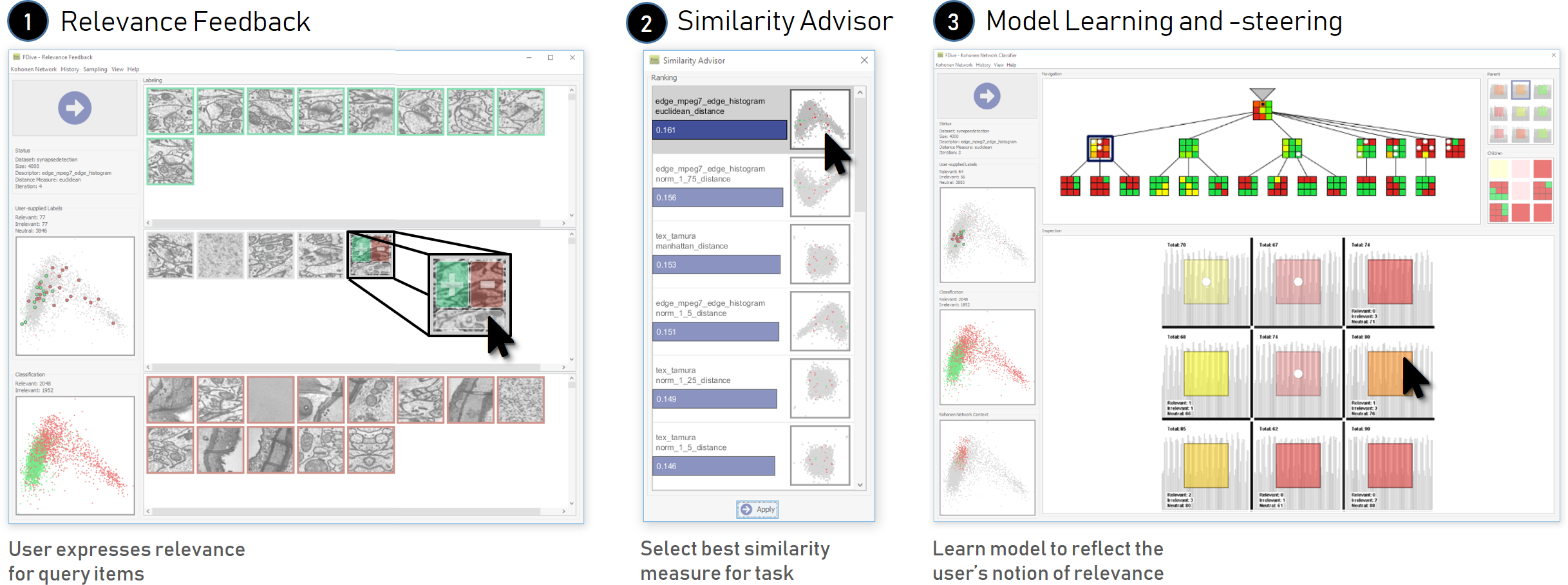} % FD and function combination
  \caption{\tool{} learns to distinguish relevant from irrelevant data through an iteratively improving classification model by learning the best-fitting feature descriptor and distance function. (1) Users express their notion of relevance by labeling a set of query items, in this case, images. (2) These labels are used to rank all similarity measures by their ability to distinguish relevant from irrelevant data. (3) The system applies the selected similarity measure to learn a Self-Organizing Map (SOM)-based relevance model. Users explore and refine the model by supplying relevance labels in uncertain data regions, especially near the decision boundaries.}
}

\abstract{The detection of interesting patterns in large high-dimensional datasets is difficult because of their dimensionality and pattern complexity. Therefore, analysts require automated support for the extraction of relevant patterns. In this paper, we present \tool{}, a visual active learning system that helps to create visually explorable relevance models, assisted by learning a pattern-based similarity. We use a small set of user-provided labels to rank similarity measures, consisting of feature descriptor and distance function combinations, by their ability to distinguish relevant from irrelevant data. Based on the best-ranked similarity measure, the system calculates an interactive Self-Organizing Map-based relevance model, which classifies data according to the cluster affiliation. It also automatically prompts further relevance feedback to improve its accuracy. Uncertain areas, especially near the decision boundaries, are highlighted and can be refined by the user. We evaluate our approach by comparison to state-of-the-art feature selection techniques and demonstrate the usefulness of our approach by a case study classifying electron microscopy images of brain cells. The results show that \tool{} enhances both the quality and understanding of relevance models and can thus lead to new insights for brain research.} % end of abstract


\begin{document}


\firstsection{Introduction}

\maketitle

%% \section{Introduction} %for journal use above \firstsection{..} instead
%\section{Introduction} % Not used here!

A  primary challenge when analyzing collected data is to distinguish relevant from irrelevant data items. Large and high-dimensional datasets are not easily analyzed, because of their size, dimensionality, and possible complex patterns. Therefore, analysts need automated support. This support is realized in the form of a relevance model that can help them to make this distinction. Its task is the retrieval of relevant data items from large high-dimensional datasets that are often associated with many types of analysis scenarios. {Similarity models are key to effective data clustering and classification.} It is crucial that the model reflects the notion of relevance as it pertains to the analysis task. More generally, when we are dealing with high-dimensional datasets, we need to automatically and adaptively assess the relevance of data items. Although analysts interact with data for analysis and exploration purposes, their primary goal is to quickly generate new insights and results. All interactions, such as labeling or relevance feedback, should be focused on yielding insights and need to be as impactful as possible.

The fully automatic creation of relevance models is non-trivial. Deep learning approaches, such as Convolutional Neural Networks (CNNs), have been applied successfully, but typically require a large number of labeled training data to distinguish relevant from irrelevant data~\cite{cnn}. Classic machine learning techniques depend on a predefined set of features and a given distance function, chosen or even designed by experts based on their experience. In most real-world scenarios, these labels do not exist and the manual assignment of labels is time consuming, tedious, and expensive. In many analysis scenarios, this is not a viable solution. Transfer learning could be an alternative solution. These methods reapply a previously learned model for a different task then that for which they were originally trained~\cite{transfer-learning}. While the idea seems intriguing, these models are unable to transfer the complex user understanding between datasets. One reason is that the problem and task definition in exploratory scenarios, particularly the pattern space, is highly specific and non-static. Users' mental model of \emph{what makes up relevance} evolves throughout an analysis, thus requiring adaptive methods for the process. Additionally, the created model needs to be understandable, explorable, and refinable in areas where it is inaccurate.

The feedback-driven view exploration pipeline by Behrisch et al.~\cite{BKSS14} was an early approach towards a relevance model-guided exploration of large multidimensional datasets. Similar to our work, the central idea is to make an arbitrary dataset accessible to users through visualizations, such as scatter plots, that can be abstracted into a set of numbers, called features. To solve real-world problems, features need to be able to express the differences between the data items concerning the analysis objective.
Features reduce the complexity of comparing data items but are limited in their ability to express all properties of a data item. The approach by Behrisch et al.~\cite{BKSS14} only uses one \emph{fixed} feature descriptor (FD), namely Scagnostics~\cite{Wilkinson2005}, limiting the set of described properties and introducing biases into the analysis process. In this work, we  tackle the  question of choosing an appropriate FD that models the given dataset, analysis domain, and analysis task. We claim that FDs alone do not express the relationship between data items. We also need a distance function that describes their relationships. Depending on the analysis scenario, other measures than the ubiquitous Euclidean distance may perform significantly better~\cite{gregor2015}, which reflects on the performance of the relevance model learning component, too. In this work, we expand on Behrisch et al.'s static decision tree model, in which exploration decisions are irreversible, with a more flexible and adaptive approach to guide the user through the data space. Our classification results and feature abstraction can be visually explained, making the quality of the model easier to capture and more trustworthy. 

In this work, we present \tool{}, a visual analytics system for the creation of relevance models. In \tool{}, we model relevance as a binary classification problem. Since the quality of the underlying classification or ranking model depends on the usefulness of the employed FDs and distance function, we introduce the concept of the \emph{Similarity Advisor} engine, which ranks FD-distance function pairs, according to their ability to distinguish relevant from irrelevant data. This removes the need for an expert choosing an FD and distance function manually. The system uses the best-ranked similarity measure for the creation of the relevance model. To learn fine-grained differences between relevant and irrelevant data, we introduce a Self-Organizing Map (SOM)-based relevance model that classifies data items according to their cluster membership. To allow the judgment of the model quality and model refinement, the SOM-based model is visually explorable and guides the user towards areas of uncertainty. We embed the \emph{Similarity Advisor} and the model learning process into an iterative framework, to allow for convergence towards the optimal similarity measure and relevance model.

We evaluate our general framework through a quantitative study comparing \tool{} to three state-of-the-art feature selection techniques, where we show that the \emph{Similarity Advisor} can outperform them {in scenarios with a low number of labels through a fast adaptation to the user's notion of relevance.} We also demonstrate \tool''s applicability and usefulness on a challenging scientific analysis task. Specifically, we consider electron microscopy images of brain cells, where a domain expert teaches the system the relevance of images depicting a neuronal synapse.

\section{Related Work}\label{sec:related-work}

In this section, we delineate \tool{} from other approaches. \tool{} is a relevance model builder, in contrast to image retrieval systems like PixSearcher \cite{Dang2014} which enables users to retrieve images through query by example. In the following, we discuss related concepts such as feature selection, visual active learning, and distance function learning. We also discuss similarities and differences in the area of model visualization and understanding.

\subsection{Feature Selection for Dimensionality Reduction}\label{subsec:feature-selection}

Feature selection algorithms typically try to approximate the usefulness of a given feature. These techniques determine a subset of relevant feature dimensions based on feature-ranking and feature-weighting~\cite{James2013, Guyon2003}. Although prior studies show how visualizations can support feature selection and optimization in 3D models~\cite{schreck_towards_2008} or exploration of chemical compounds~\cite{Strobelt2012, bremm}, the feature evaluation procedure is reoccurring and potentially exhausting for the user. Thus, we decided to use two purely automatic statistical feature selection algorithms in the evaluation of \tool{}. First, \emph{ReliefF}~\cite{Koprinska2009, Wang2016} is a state-of-the-art extension of the \emph{Relief} algorithm for multi-class problems~\cite{mehri}. It ranks features based on how well they distinguish an instance from its $k$-nearest neighbors. If a neighbor is from a different class, the weights of features that separate both instances are increased, and all others are decreased accordingly. In case the neighbor is from the same class, the weights of features that separate both instances are decreased, and all others increased. Second, \emph{Linear Ranking Ensembles} combine multiple ranking classifiers, such as the \emph{Recursive Elimination Support Vector Machine (SVM)}, into one ranking ensemble. They are, thus, more stable than other approaches~\cite{Saeys2008}. \emph{Recursive Elimination SVMs} iteratively reduce the feature dimensions size using linear SVMs~\cite{Lin2012}. Attributes are ranked, and the worst performing dimension is removed. This process, including the SVM training, continues until only one feature dimension remains. 

The quality of a feature selection depends on the number of available labels and is computationally expensive in scenarios that require continuous reevaluation. With \tool{}, we provide a solution for this scenario by keeping the feature descriptions while ranking a set of similarity measures, consisting of an FD and a distance function combination, based on how well it separates relevant from irrelevant data. We embed this technique in an iterative process, allowing for an adaptation to the best-suited similarity measure.

\subsection{Visual Active and Interactive Machine Learning}\label{subsec:active-learning}

In a visual active learning (AL) system, users are provided with auxiliary information about the learning process and model state, specifically decision boundaries of the classification model, query choice, and learned instances. Bernard et al.~\cite{bernard-1} present a visual AL method to assess the well-being of prostate cancer patients from the patient's history, describing interesting biological and therapy events. The tool suggests a set of candidates to label, as well as allowing for the visual verification of the validity of learned instances. Heimerl et al.~\cite{heimerl} present a visual AL system as an SVM classifier for text. The tool supports the visualization of the decision boundary, including instances on it, and user-based instance selection for labeling. Eaton et al.~\cite{eaton} adjust the underlying data space by describing it with manifold geometry, allowing users to label data items, serving as control points leading to improved learning performance.

In contrast to AL, the sample selection in interactive machine learning (IML) is driven by the user. Dudley et al.~\cite{Dudley2018} describe a general approach to interface design for IML providing an overview of challenges and common guiding principles. Arendt et al.~\cite{Arendt2019} present an IML interface with model feedback after every interaction by updating the items shown for each class. The users can drag misplace data items to the appropriate class and, if needed, create a new one. Both actions update and improve the model.

\tool{} is a visual active learning system that learns a relevance model based on the user's notion of relevance. We propose a SOM-based model, which is interactively explorable, guiding the user to areas of uncertainty and decision boundaries. The model creation and inspection are combined in an iterative workflow that allows the user to observe and judge model change, leading to a more understandable relevance model and learning process.

\subsection{Distance Function Learning}\label{subsec:distance-function-learning}

Another requirement to represent the relationship of data items is a distance function. A distance function can include a feature weighting. The Mahalanobis metric~\cite{mahalanobis} measures the standardized distance of a data point to the estimated mean of its population. Relevant Component Analysis~\cite{mm-rca} uses a parameterized Mahalanobis distance. This technique adapts the feature space by assigning large weights to relevant dimensions and low weights to irrelevant dimensions through equivalency constraints, describing the similarity of data items. As opposed to purely algorithmic approaches, there are also visual and interactive approaches to the generation of suitable distance functions. Brown et al.~\cite{dis-function} learn a distance function from a 2-dimensional projection of the data space where the user drags the data point to the desired position, thus describing similarity relations. The underlying distance function is updated accordingly by the adaptation of feature weights. The work by Gleicher~\cite{gleicher} demonstrates the learning of multiple distance functions, each describing the relationship of the data based on different features, capable of describing abstract concepts, such as socio-cultural properties of cities. Fogarty et al.~\cite{cue-flick} present an image retrieval system that determines the weights of a distance metric based on user-supplied feedback to learn concepts.

In contrast, \tool{} unifies many concepts mentioned above. It ranks arbitrary feature descriptors and similarity measure combinations by their ability to discriminate relevant from irrelevant data. \tool{} removes the limitation on a pre-defined set of features through the comparison of multiple FDs describing a diverse set of data properties. Also, a set of similarity coefficients is used, thus removing the limitation of a single similarity coefficient or feature weighting. This makes \tool{} a generalized relevance model builder for different types of data.

\subsection{Model Visualization and Understanding}\label{subsec:model-visualization}

Visual Analytics (VA) aims to provide the analyst with visual user interfaces that tightly integrate automatically obtained results with user feedback~\cite{keim_mastering_2010}. The knowledge generation model~\cite{Sacha2014} describes an iterative process of exploration and verification activities of both human and machine. Results are presented visually to analysts, who interpret obtained patterns and provide feedback to steer the exploration process or form and refine hypotheses. The understanding and interpretation of machine-learned models is key for the effective incorporation of user feedback in such scenarios. Several prior works have studied model visualizations and interactions. BaobabView~\cite{Elzen2011} presents a model where the structure of a decision tree is augmented with data distributions and data flows. Liu and Salvendy~\cite{Liu2007a} and Ankerst et al.~\cite{Ankerst2000} use icicle plots~\cite{Kruskal1983,kleiner81} to visualize decision trees. Visual interactive approaches for cluster evaluation and understanding were presented by Nam et al.\ for general high-dimensional data~\cite{Nam2007} and by Ruppert et al.~\cite{Ruppert2017} for the clustering of text documents.  Sacha et al.\ present SOMFlow~\cite{Sacha2018}, an exploration system that uses Self-Organizing Maps (SOM) to guide the user through an iterative cluster refinement task, leveraging the proximity-preserving property of SOMs~\cite{Ultsch1999, Bernard2012}  for clustering and data partitioning tasks.

In a model creation task, the user needs to be guided towards areas of high uncertainty. Thus \tool{} steers the data exploration to specific parts of the model, such as the decision boundaries. The SOM-based model of \tool{} is capable of providing the necessary information about uncertain areas and automatic refinement.

\section{Similarity Advised Model Learning}\label{sec:fdive-pipeline}

\begin{figure}[b]
    \centering
    \includegraphics[width=\linewidth]{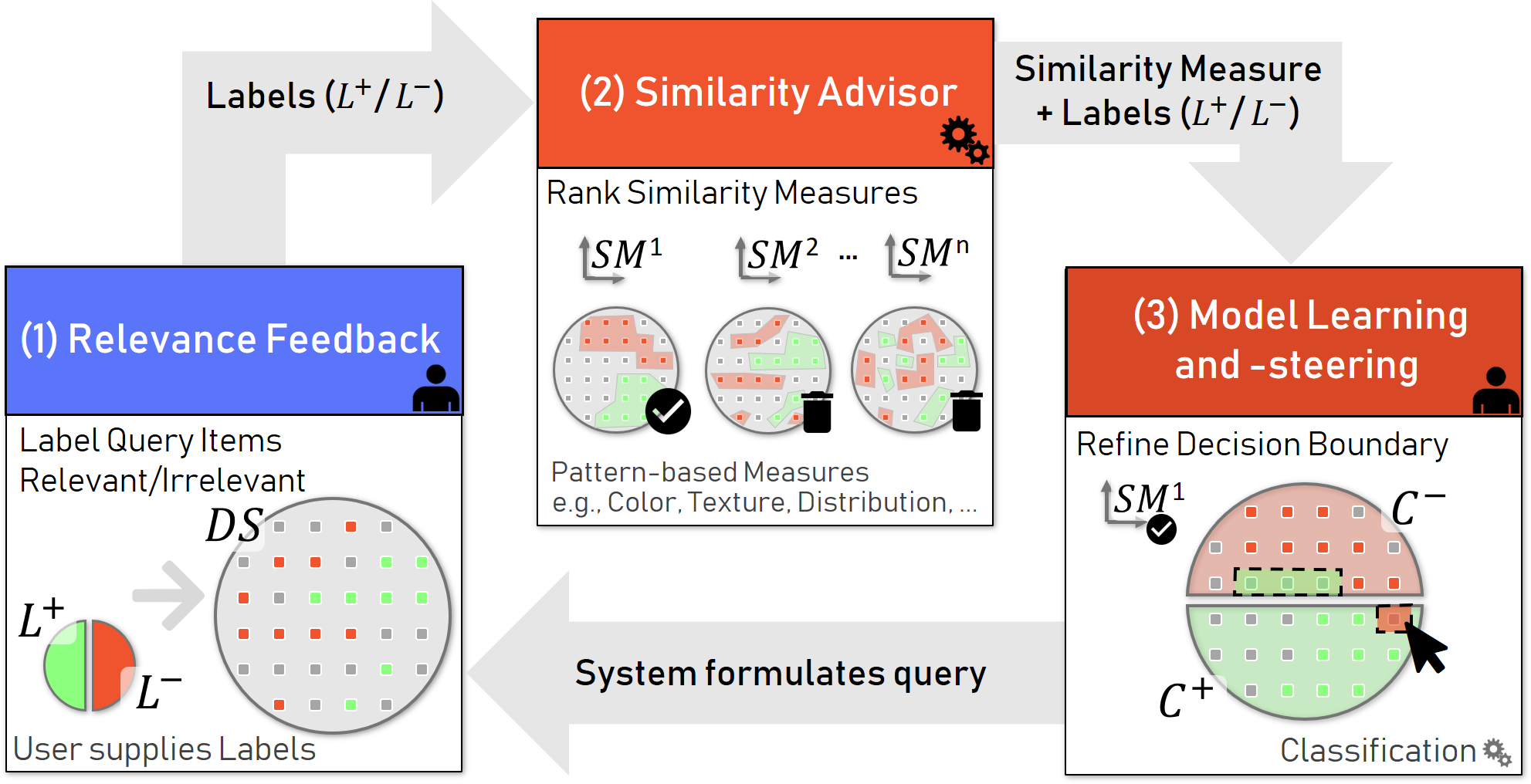}
    \caption{(1) Users label query items as relevant or irrelevant and therein express their notion of relevance. (2) This selection is used to automatically determine the best-fitting similarity measure, which distinguishes relevant from irrelevant data. (3) The system adapts the model using the relevance labels and similarity measure. The model is explorable and refinable by the users, to improve its accuracy.}
    \label{fig:detail-workflow}
\end{figure}

The key idea of our approach is to iteratively and interactively create relevance models, where a useful feature description is unknown, and no or only few labels are available. Our proposed \emph{Similarity Advisor} allows approaching the question which feature descriptor and similarity measure combination is useful to distinguish relevant from irrelevant data items. In a scenario where labels are sparse, the quantitative validation of classification models with performance measures is inexpressive. Thus, there is a need for techniques that allow for model assessment without test data. Classifiers, such as SVMs, have been used in visual active learning approaches~\cite{heimerl}. However, the representation of the data space created by SVMs does not allow the user to judge the quality of a classifier visually. Decision trees are more intuitively interpretable.

We propose a SOM-based classification model which is embedded in an iterative workflow to allow for observable learning steps. In each step, the model is explorable and refinable to judge and improve its quality. Both, the \emph{Similarity Advisor} and the SOM-based classification model constitute \tool{}, a generalized model builder. In the following, we provide an introduction to SOMs.

\smallskip

\noindent\textbf{Self-Organizing Maps:} \tool{} relies on a \emph{neural network architecture}, called Self-Organizing-Map (SOM) or Kohonen Network. SOMs are the basic building block of our relevance model and are one of the classical neural network structures, created by Kohonen to derive topologically coherent feature maps~\cite{Kohonen1982}. SOMs can be visualized as a grid of cells representing the neurons of the network. The cells contain prototype vectors representing data clusters. In the learning phase of the network, the most similar prototype vector (best-matching-unit) to the training input is identified and adjusted towards the input vector. Spatially close neighbors are also adapted, depending on a learning rate and radius parameter. The latter gives rise to the self-organization property of the map. The final result is a topology, where data items are clustered. Clusters can consist of single or multiple cells, and cluster similarity can be captured by spatial proximity of clusters on the SOM grid~\cite{Bernard2012, Sacha2018}.

\smallskip

We extend this algorithm into a tree-like classifier to allow for the representation of fine-grained similarity differences. This concept is based on the idea that items can ``flow'' from a parent SOM node into a child SOM for further analysis, as presented by Sacha et al.~\cite{Sacha2018}. In our work, we extend this idea to create a classification model that \emph{automatically} partitions the high-dimensional data space into relevant and irrelevant data item clusters. We will detail this approach in \autoref{sec:som-model}. We use an interactive SOM visualization to allow for the visual inspection of the currently learned model, e.g., where groups of relevant or irrelevant data elements are located, and how well decision boundaries can distinguish known groups.

\subsection{Workflow for Iterative Relevance Model Learning}\label{sec:tool-pipeline-and-workflow}

\tool{} is inspired by the feedback-driven interactive exploration tool by Behrisch et al.~\cite{BKSS14}, which propose an iterative and FD-based exploration framework. A central principle is to represent an arbitrary dataset with the help of visualizations to make it accessible for an analyst. This visualization needs to be translated into a language understood by a computer, which uses this \emph{proxy} to guide through the information- and pattern space which is achieved by a single fixed FD introducing bias into the analysis process.

We expand this body of work by changing the focus from an exploratory approach to a model building technique. The validation of relevance models, though, is a challenging task, due to the following reasons. We need to define a useful definition of similarity, but a metric for separating classes can only be determined during the learning process. What is needed are flexible and adaptive strategies for determining a useful metric defining similarity. \tool{} allows for arbitrary data modeling through the \emph{Similarity Advisor}, which ranks a set of FDs and distance functions by their usefulness concerning the current analysis domain and dataset properties. The FD, representing the data modeling, is subsequently used to create a relevance model. Additionally, the model needs to be explorable and refinable to convince an analyst of its usefulness and accuracy. 

In \tool{}, we leverage an iterative workflow to continuously revalidate the similarity measure and improve the relevance model. In the following, we describe each iteration step and its impact. \autoref{fig:detail-workflow} shows each step accordingly.

\smallskip
\noindent\myboxrel{\textbf{(1)}} \textbf{Relevance Feedback:}
The system prompts the user to label a subset of data items of the dataset ($DS$) as relevant or irrelevant, representing relevance as a binary classification problem. Those data items labeled as relevant are referred to as $\mathpzc{L}^+$ and all labeled as irrelevant as $\mathpzc{L}^-$. Unlabeled data items are considered neutral. In the first iterations, this step is replaced by a query generated through a representative data sample. In all following iterations, the query is determined by the SOM-based model. \tool{} supports the user by visual feedback allowing the validity assessment of a currently used similarity measure and classification through visual feedback. This step is described in detail in \autoref{sec:annotation-view}.

\smallskip
\noindent\myboxfs{\textbf{(2)}} \textbf{Similarity Advisor:} The system evaluates all possible pair-wise combinations of FDs and distance functions by their ability to separate relevant ($\mathpzc{L}^+$) from irrelevant ($\mathpzc{L}^-$) data items. A ranking shows the evaluation result, giving an intuition about the similarity measures. The user can follow the recommendation or choose a different similarity measure. The system uses the FD and distance function for the creation of the relevance model. We describe the algorithmic background of the \emph{Similarity Advisor} in \autoref{sec:similarity-advisor}.

\smallskip
\noindent\myboxml{\textbf{(3)}} \textbf{Model Learning and Steering:} The system creates a classification model based on the selected similarity measure and available labeled data ($\mathpzc{L}^+$ and $\mathpzc{L}^-$). The model can be explored to asses its properties and viability for its classification task. The classification result is referred to with $\mathpzc{C}^+$ describing all data items classified as relevant and all irrelevant as $\mathpzc{C}^-$. The SOM-model creation and interactions are described in \autoref{sec:som-model}. Subsequently, the system determines a set of query items which are labeled by the user in the first step of the next iteration.

\smallskip

\noindent In the following, we  describe the design, user interaction and algorithmic support in \tool{}.

\section{Context-aware Relevance Feedback}\label{sec:annotation-view}

Data labeling is the first and reoccurring step in our relevance model learning process from~\autoref{sec:tool-pipeline-and-workflow}. During start-up, this essential bootstrapping step helps us to form a decision basis for the subsequent application of our \emph{Similarity Advisor}. Throughout the learning process the classifier queries relevance labels through this interface to improve its accuracy. We describe this step of \tool{} in \autoref{sec:som-model}.

\begin{figure}[tb]
\centering
\includegraphics[width=\linewidth]{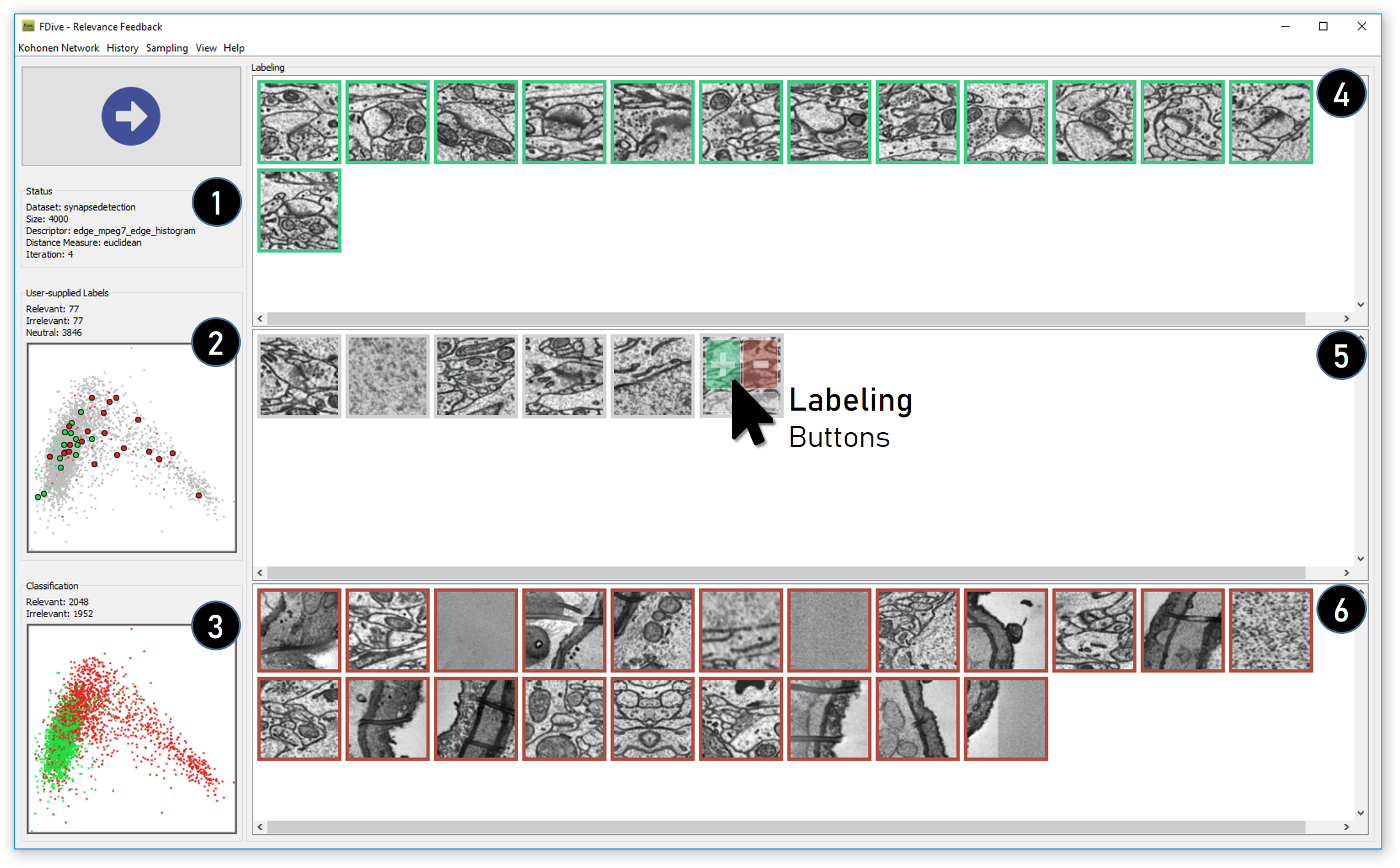}
\caption{Context-aware Relevance Feedback: (1) Status display showing the current analysis state. (2) Scatter plot highlighting newly labeled data. (3) Scatter plot of the current classification result. Both allow judging the impact of new labels. (5) Queried neutral data items. Data items labeled as relevant (4) and irrelevant (6).}
\label{fig:labeling}
\end{figure}

\subsection{Relevance Feedback of Representatives}

We sample data items in the first iteration for an initial user labeling. The sampling method can be chosen from the following options: Minimum-Maximum-, Quantile Sampling, Normal-, Stratified Normal Bootstrapping, Normal- or Stratified Subsampling~\cite{BKSS14}. In all following iterations, the request for labeling is determined by the relevance model, in our case a Self-Organizing Map-based model (\autoref{sec:som-model}). The user can apply three types of labels: relevant, irrelevant and neutral. While the relevant and irrelevant labels express a user preference and have an impact on all steps of \tool{}, neutral represents an uncertain item. The model may prompt a label for the given element at a later iteration. The user labels a subset of displayed data items by clicking on the mouse-over menu or using a keyboard-shortcut. For visual clarity, all elements are assigned to specific panels (relevant, neutral, irrelevant, from top to bottom in \autoref{fig:labeling} (3-5), according to their label type, which also allows comparing items with the same relevance label.

\subsection{Visual Assessment of Labeling Impact}

A status display (\autoref{fig:labeling} (1)) provides information about the current analysis state, such as the current FD and distance function, the number of supplied relevant and irrelevant labels, as well as the number of remaining neutral items. A scatter plot (\autoref{fig:labeling} (2)) of the dataset using the currently chosen FD and distance function depicts the possible impact of new labels when compared to the projection of the classification result (\autoref{fig:labeling} (3)). We create both 2D projections with multi-dimensional scaling (MDS). MDS projects the data in a distance-preserving way without the need for additional parameters. The annotation view is also used to refine the labels in the SOM-based model and explore elements assigned to a SOM-neuron (\autoref{sec:som-model}). Chegini et al. explored the idea of showing the classification result in a scatter plot~\cite{chegini}, while the visual feedback on data labeling was evaluated by Bernard et al.~\cite{Bernard2018ComparingVL}. Combining both approaches allows assessing the impact of newly assigned labels in a natural form. The comparison of both scatter plots shows the effect of new labels, e.g., a relevant label in an area of irrelevant classifications hints at an incomplete reflection of the user's notion of relevance, a matching label hints at a convergence.

\section{Assessing Pattern-based Similarity Measures}\label{sec:similarity-advisor}

\begin{figure*}[tb]
\centering
\includegraphics[width=0.9\linewidth]{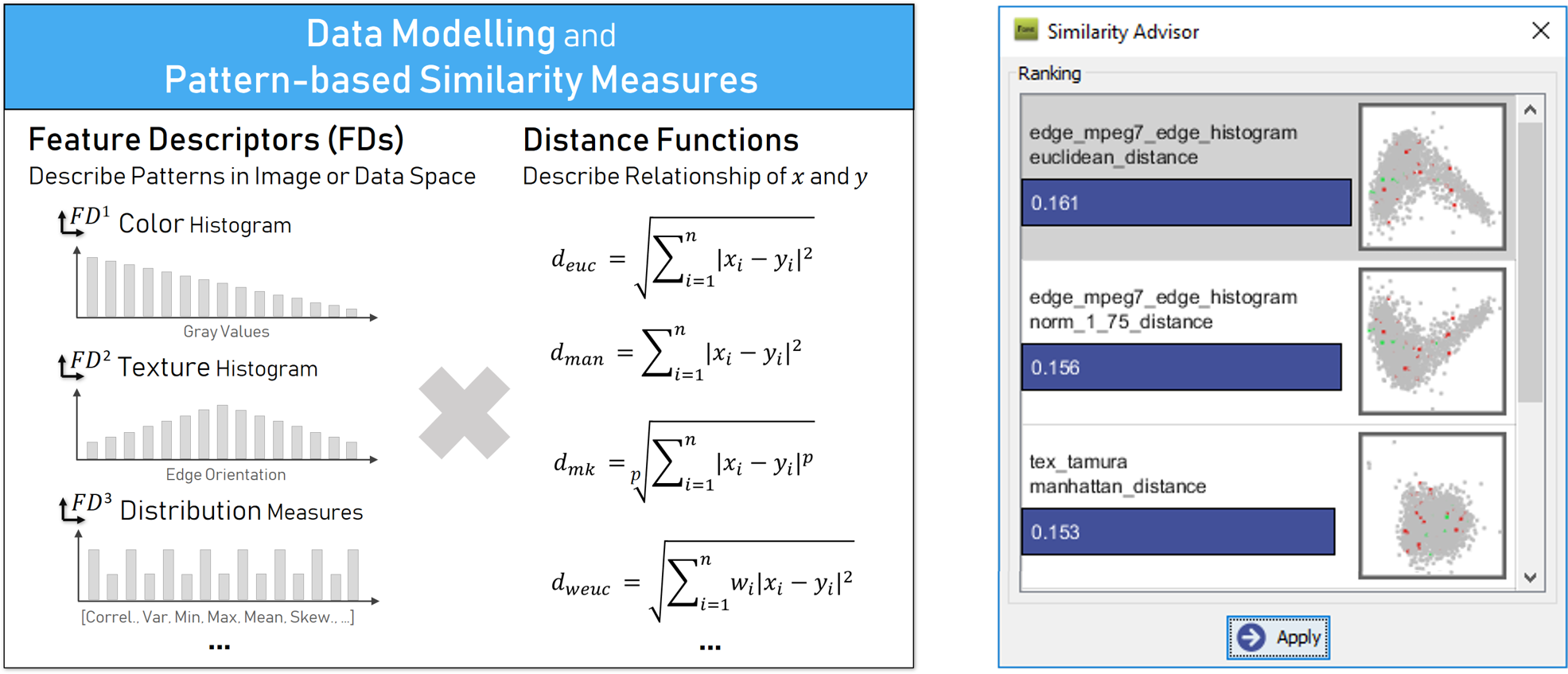}
\hfill
\caption{Left -- The \emph{Similarity Advisor} uses a set of FDs and distance functions. FDs model the data based on perceptible patterns in the data or image space. Distance functions describe the relationship between two points in the FD  space. In \tool{}, we consider all pair-wise combinations as potentially useful measures. We call a combination of an FD and a distance function a \emph{pattern-based similarity measure}. Right -- The \emph{Similarity Advisor} ranks all pair-wise combinations of FDs and distance functions according to their ability to distinguish relevant from irrelevant data. A bar indicates the score and a scatter plot shows the topology of implied data distribution {allowing users to judge its usefulness.}}
\label{fig:similarity-advisor}
\end{figure*}

The goal of the \emph{Similarity Advisor} is to select the most expressive FD and distance function combination from a predefined set of FDs and distance measures to improve the relevance model creation. We claim that a combination of FD and distance measure can define a \emph{pattern-based similarity measure}. To describe the discriminative ability, we need a \emph{quality metric} that reflects the similarity measure's ability to distinguish between our relevant and irrelevant items. We consider a useful similarity measure one that maximizes the distance between both sets $\mathpzc{L}^+$ and $\mathpzc{L}^-$. We considered other quality metrics, such as metrics that measure distances between elements of a cluster, but found them lacking in performance. We propose the \emph{Similarity Advisor} for the selection of a suitable distance metric; this includes the choice of an FD and a distance function. For this, we require a set of diverse FDs. We use various FDs from the Image Processing and Computer Vision Community because these algorithms are designed to match the human perceptual system.

\medskip

In essence, the application scenario determines the usefulness of a feature description and distance function. However, the selection of a useful distance function is hard. Thus, we introduce the concept of continuously evaluating a set of \emph{pattern-based similarity measures} for their applicability to the current analysis task, allowing for the convergence to the most useful one. To describe the algorithmic basis of the \emph{Similarity Advisor}, we define all relevant terms.

\bigskip

\noindent\textbf{Feature Descriptor (FD):} FDs are modeling specific characteristics of a data item. Examples for low-level FDs are color histogram descriptors, modeling the color distribution, or edge histograms describing edge orientations of an image~\cite{Rui1999}. Low-level FDs are typically inexpensive to compute and may work robustly. Depending on the type of data at hand, many FDs are applicable. Mathematically, an FD can be described as a function $FD: DS \to \mathbb{R}^n$, where $DS$ denotes the dataset and $\mathbb{R}^n$ the implied vector space. The dimensionality $n$ depends on the FD. \autoref{tab:feature-descriptors} lists all FDs used by \tool{}. These FDs describe a variety of different image features, such as color, layout, structure, and shape~\cite{Behrisch2016a}.

\medskip

\noindent\textbf{Feature Vector (FV):} An FV is an instantiation of an FD for a specific data item. An FV contains one or multiple components, called feature dimensions or features. A feature vector $FD(x) \in \mathbb{R}^n$ represents a description of a data item $x \in DS$, w.r.t. the properties described by the applied FD. \\

\smallskip

\noindent\textbf{Feature Space (FS):} A feature space describes the set of all feature vectors created by an individual feature descriptor. Additionally, a feature descriptor implies a vector space, called feature space. Thus, each feature descriptor has an associated vector space.

\smallskip

\begin{table}[bt]
\setlength{\tabcolsep}{5pt}
\centering
{\small
\begin{tabular}{p{.46\linewidth}p{.46\linewidth}}
\toprule
\textbf{Color} & \textbf{Color Layout} \\
\toprule
\textsc{Auto Color Correlogram}~\cite{609412} & \textsc{Cedd}~\cite{Chatzichristofis2008} \\
\textsc{Fuzzy Histogram}~\cite{Han2002} & \textsc{Fcth}~\cite{Chatzichristofis2008fcth} \\
\textsc{Fuzzy Opponent Histogram}~\cite{Sande2010} & \textsc{Jcd}~\cite{Chatzichristofis2008} \\
\textsc{Global Color Histogram}~\cite{Rui1999} & \textsc{Luminance Layout}~\cite{Lux2008} \\
\textsc{Opponent Histogram}~\cite{Sande2010} & \textsc{MPEG7 Color Layout}~\cite{Kasutani2001} \\
\toprule
\textbf{Edge} & \textbf{Structure} \\
\toprule
\textsc{Edgehist}~\cite{Behrisch2016a} & \textsc{JPEG Coefficient Histogram}~\cite{Lux2008} \\
\textsc{MPEG7 Edge Histogram}~\cite{Morse2000} & \textsc{Phog}~\cite{Bosch07} \\
\textsc{Hough}~\cite{Hough1962} & \textsc{Profile}~\cite{Behrisch2016a} \\
\toprule
\textbf{Texture} & \textbf{Other} \\
\toprule
\textsc{Gabor}~\cite{Lux2008} & \textsc{Blocks}~\cite{Behrisch2016a} \\
\textsc{Haralick}~\cite{Haralick1973} & \textsc{Compactness}~\cite{Morse2000} \\
\textsc{Local Binary Pattern}~\cite{Heikklae2006} & \textsc{Magnostics}~\cite{Behrisch2016a} \\
\textsc{Tamura}~\cite{Tamura1978} & \textsc{Statistical Noise}~\cite{Behrisch2016a}\\
\bottomrule
\\
\end{tabular}
\caption{\tool{} uses 24 feature descriptors. These FDs describe a variety of different image features, such as color, layout, structure and shape~\cite{Behrisch2016a} allowing for a description of a diverse set of properties.}
\label{tab:feature-descriptors}
\vspace*{-1.5em}
}
\end{table}

\noindent\textbf{Pattern-based Similarity Measure:} We define a \emph{pattern-based similarity measure} as a combination of one feature descriptor and a single distance function (\autoref{fig:similarity-advisor} (left)). The \emph{Similarity Advisor} evaluates the usefulness all possible combinations of an FD and a distance function in their ability to separate the clusters of relevant ($L^+$) and irrelevant ($L^-$) data items.   

\smallskip

In \tool{}, we use a set of norms as distance functions because the SOM learning algorithm requires a similarity measure that can describe a vector space allowing for an adaptation of the cluster prototypes ``towards'' an input vector. \tool{} uses the following norms: Euclidean $L^2$, Manhattan $L^1$, $L^{1.25}$-norm, $L^{1.5}$-norm and $L^{1.75}$-norm, which are all $L^p$-norms with $||x||^p=(\sum^d_{i=1}|x_i|^p)^{1/p}$ and the implied metric $d(x,y)=||x-y||$ as a similarity measure.

\subsection{Comparability of Pattern-based Similarity Measures}

Every FD describes a different set of data properties by mapping a data item to a vector representation. To derive useful similarity relations, we need to use a distance function that applies to the vector. We limit ourselves to $L^p$-norms. However, this approach is extendable to other distance functions and similarity coefficients, including those which do not satisfy the metric axioms.

We leverage the definition of normed vector spaces, which is defined as $(V, ||\cdot||)$ where $V$ is a vector space and $||\cdot||$ a norm on $V$. We use this definition and apply it to the combination of an FD and its FS along with an $L^p$-norm with $p \in [1, \infty)$. Throughout this paper, the term distance function refers to the induced metric $d(x, y) = ||x - y||$. In \tool{}, we define a \emph{pattern-based dissimilarity measure}, a combination of a single FD and a distance function, formally as $dist^{FD}_d: (x, y) \to [0, \infty)$ with $dist^{FD}_d(x,y) = d(FD(x), FD(y))$ and $x, y \in DS$ data items of the dataset. 

We apply a non-standard normalization to transfer a feature space $FS$ and the associated norm into a comparable format. To achieve this outcome, we center the set of all feature vectors $x \in FS$ on the origin, such that the center of each dimension range is located at the origin. This translation does not change vector distances. For this we create a translation vector $t \in \mathbb{R}^n$.
The components of $t$ are defined for each dimension $i$ as
\begin{equation}
    t_i = 0.5 \cdot (max_{v \in FS}(v_i) + min_{v \in FS}(v_i))
\end{equation}

With this, we can formalize the necessary normalization step to transform the feature space into a comparable state as described by the following function.
\begin{equation}
    normalize(x) = (x - t) / max_{v \in FS}(||v - t||) \text{ with } x \in FS 
\end{equation}

The normalization needs to be performed for all elements $x$ of feature space to convert it into a comparable format. This normalization can be implemented with a complexity of $O(N \cdot M)$ for the full dataset of size $N$ and $M$ \emph{pattern-based similarity measures} implied by the similarity measures, leveraging the mathematical definition of a norm. In essence, this transformation translates all vectors such that the center of each dimension range is located at the origin and scales all vectors such that $||x|| \in [0,1]$ for all vectors $x$, while preserving relative distances between all vectors according to the norm. This normalization allows us to compare the different topologies created by different feature descriptor and norm combinations.

This approach can extend to non-norm similarity coefficients, under the following implications. (1) Ideally, the subsequently applied classification model is compatible with the similarity coefficient, e.g., Self-Organizing Maps require a norm as an internal distance function since prototype vectors need to be updated ``towards'' an input vector. (2) With non-norm similarity coefficients, the following non-standard normalization needs to be performed. Non-norm similarity coefficients define the difference purely by the distance of data items. This requires the normalization of the full distance matrix of the feature space. This leads to a significant complexity increase since all pair-wise distances need to be computed in $O(N^2 \cdot M)$.

\subsection{Quality Metrics for Pattern-based Similarity Measures}

\begin{figure}[tb]
\centering
\includegraphics[width=0.82\linewidth]{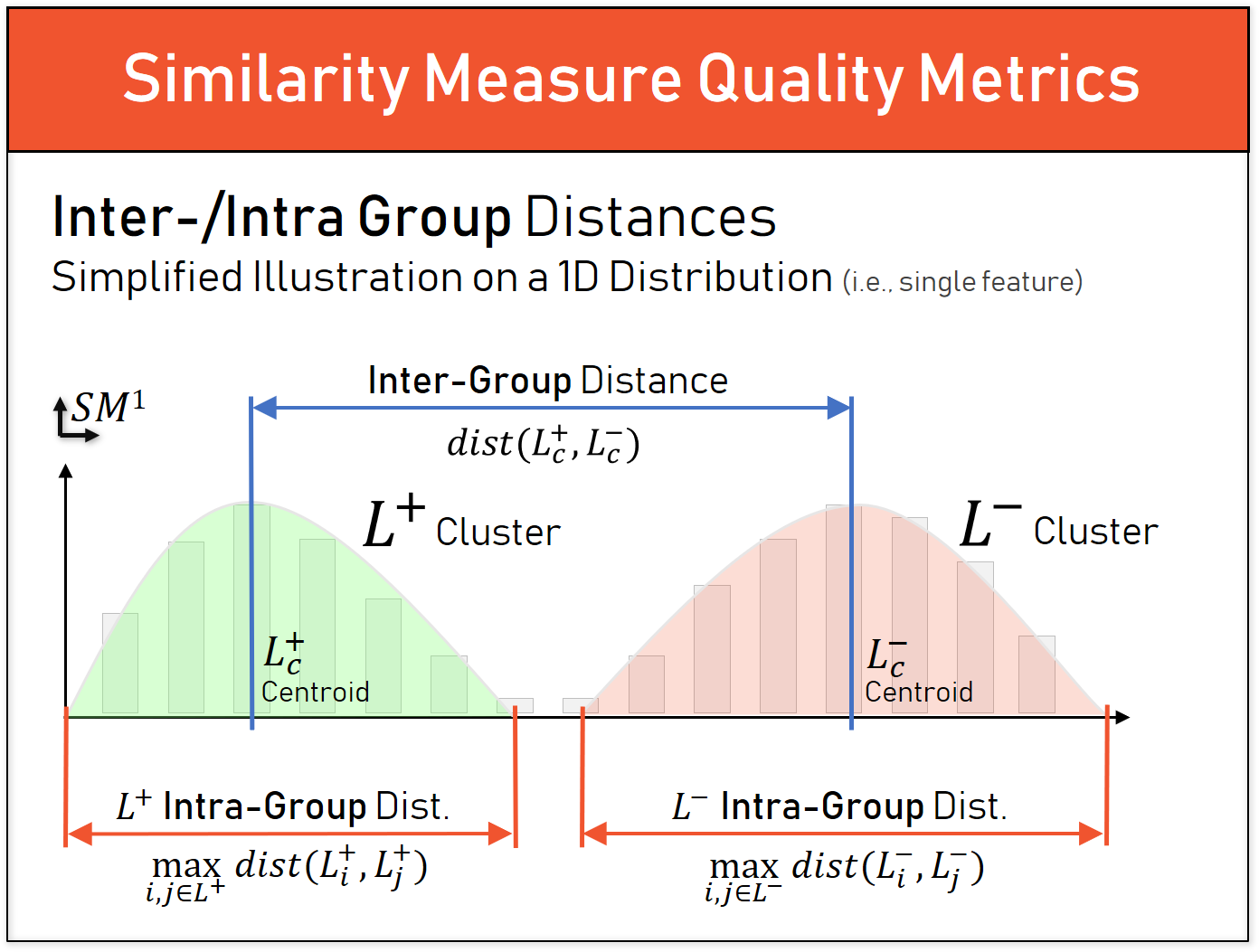}
\caption{We propose two quality metrics to evaluate similarity measures. \emph{Inter-Group-Distance} describes the distance between the centroids of the relevant and irrelevant data, measuring how well a similarity measure separates both groups. The \emph{Intra-Group-Distance} is defined as the maximum distance in the relevant or irrelevant data, measuring whether a similarity measure describes elements of the same group to be dissimilar.}
\label{fig:algo-distances-nested-som}
\end{figure}

In this section, we discuss a set of heuristic quality metrics that we designed to estimate the applicability of a similarity measure for a given analysis task. All quality metrics are calculated based on the transformed features space and the associated distance function, according to the previous section. We measure two concepts, \emph{Inter-Group-Distance}, and \emph{Intra-Group-Distance}. A group is defined as a set data items sharing an identical label, i.e., relevant or irrelevant. Thus one group is formed by all elements in $\mathpzc{L}^+$ and another by $\mathpzc{L}^+$. An intuition is given in~\autoref{fig:algo-distances-nested-som}.

\smallskip

\noindent\textbf{Inter-Group-Distance} measures the similarity of the groups, by calculating synthetic centroids of $\mathpzc{L}^+$ and $\mathpzc{L}^-$, and subsequently determining the distance between both centroids or short $Q_{inter}(\mathpzc{L}^+, \mathpzc{L}^-) = dist(\mathpzc{L}^+_c, \mathpzc{L}^-_c)$. A large \emph{Inter-Group-Distance} is highly desirable.

\smallskip

\noindent\textbf{Intra-Group-Distance} measures the maximum distance between distinct elements one of label, i.e. $\mathpzc{L}^+$ and $\mathpzc{L}^-$. Thus, we can say $Q_{intra}(\mathpzc{L}) = max_{i, j \in \mathpzc{L}}(dist(\mathpzc{L}_i, \mathpzc{L}_j))$, where $i \neq j$. We will apply the above heuristic for every dissimilarity measure.

\smallskip

We experimented with different combinations of \emph{Inter-} and \emph{Intra-Group-Distance} and variants also involving mean and median values instead of the maximum for the \emph{Intra-group-distance}. We also combined both measures into $Q_{comb}(\mathpzc{L}^+, \mathpzc{L}^-) = Q_{inter}(\mathpzc{L}^+, \mathpzc{L}^-) - w \cdot (Q_{intra}(\mathpzc{L}^+) + Q_{intra}(\mathpzc{L}^-))$, with a weighting $w$. In general, we found that the \emph{Inter-Group-Distance} performed the best on its own, i.e., with $w = 0$. 

Other metrics in the context of internal cluster quality metrics use similar notions to \emph{Inter-} and \emph{Intra-Group-Distance}. Cutting et al.~\cite{Cutting1992} describe internal cluster metrics such as the cluster self-similarity defined as the average distance of all cluster members or the average distance of all cluster members to the centroid. We found that this measure did not describe the group separation very well since the ideal case describes a cluster concentrated on a small region. This case rarely occurs in real-world scenarios, without all points of both $L^+$ and $L^-$ clusters being concentrated at the same location. We looked at internal cluster quality metrics such as the Dunn  Index~\cite{Dunn1973} which measures the ratio of minimum cluster distance to the maximum cluster extent. Another measure is the Davies-Bouldin index~\cite{Davies1979} describing the sum of cluster extents to the centroid distances. Both approaches include the notion similar to the \emph{Intra-Group-Distance}.
We found that both measures were sensitive to outliers and thus where not as useful as the \emph{Inter-Group-Distance} heuristic.

We use and suggest the \emph{Inter-Group-Distance} on its own in all applications and evaluations of \tool{}. This distance-based score is used to rank the set of similarity in descending order, as shown in~\autoref{fig:similarity-advisor}~(right). The \emph{Similarity Advisor} shows the score as bar. Additionally, we display the topology of the associated features space. Labeled data items are highlighted, allowing users to verify the separation of relevant and irrelevant data items. With the \emph{Inter-Group-Distance} we found a heuristic that is intuitive, easy to calculate and performs well, as we will show in \autoref{subsec:quant-evaluation}.

\section{Learning Relevance of Data Points with Self-Organizing Maps}\label{sec:som-model}

\tool{} features a SOM-based classifier, which is used to classify data items by their assignment to a SOM-neuron, and to learn decision planes in the high-dimensional space discriminating \ASplus and \ASminus. We introduce a set of visual encodings to guide the user to potentially interesting data subsets, or regions of classifier uncertainty.

\begin{figure}[tb]
\centering
\includegraphics[width=\linewidth]{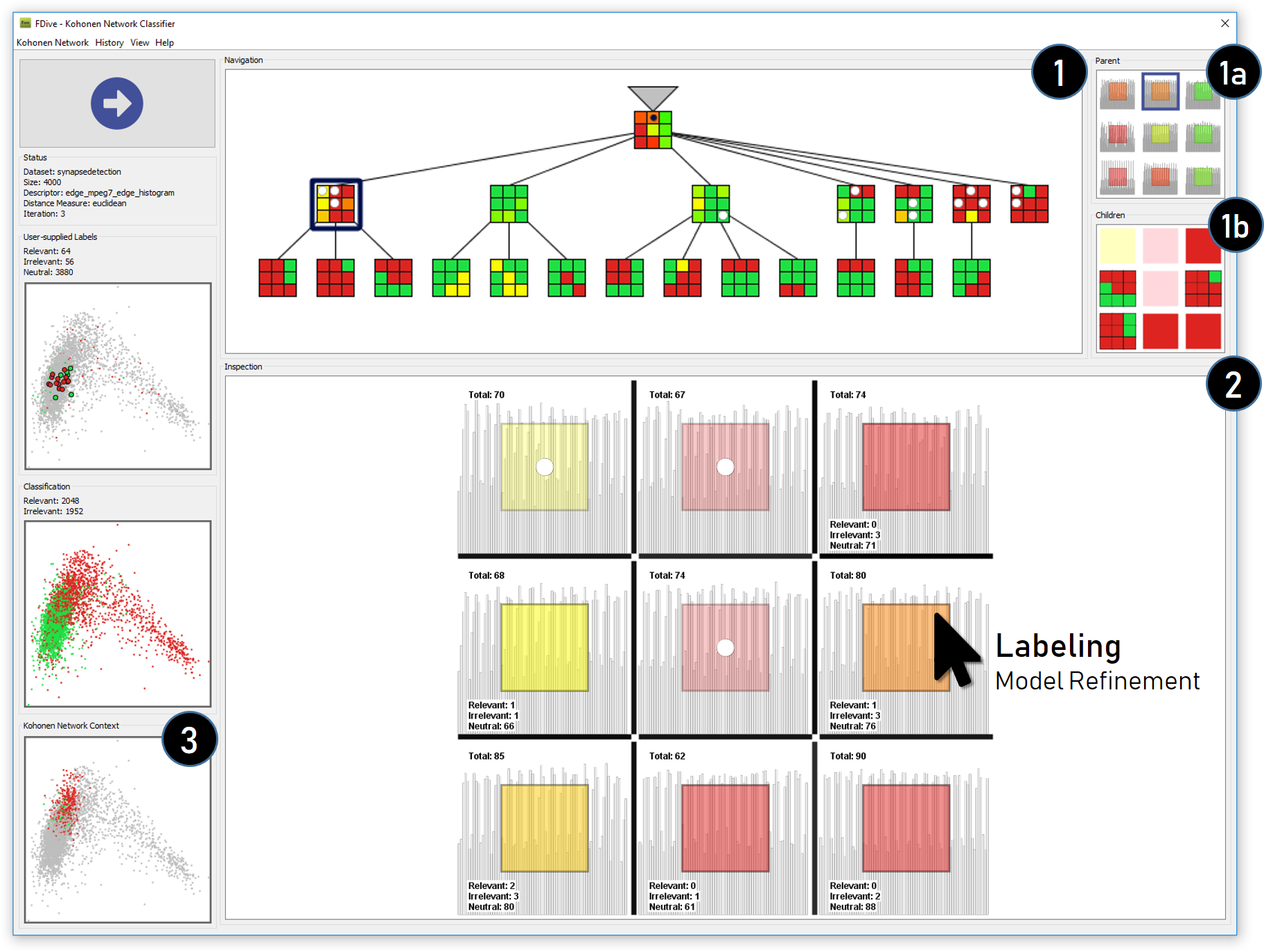}
\caption{Visual Exploration of SOM Model: (1) Classifier tree. (1a) Parent of the currently observed SOM. (1b) Children of the current SOM. (2) Detailed SOM Display. (3) Scatter plot highlighting data of the SOM node.}
\label{fig:som-classifier}
\end{figure}

\subsection{SOM as Visual Classifier}

SOMs cluster similar items in cells, which provides users with an intuition about the classification process. SOMs preserve distance relations between cells allowing for orientation in the data space. The tree-structure and SOM cell exploration allow for a drill-down from the data space to clusters and individual data items. SOM cells are arranged in a grid which is directly visualizable, which also applies and our tree-like classifier model. Additionally, our SOM classifier conveys areas of uncertain classifications by highlighting cells with mixed labeling and cells with a low amount of labeled data items. Additional labels improve the classification. Labels can be added in those specific areas. The grid size is a user parameter, and $3\times3$ is the default setting.

We use Self-Organizing Maps as a basis for our model because it is visually explorable; it partitions the feature space and the data space, which provides the user with analyzable chunks. The supplied relevance labels and the selected dissimilarity definition are used to calculate a SOM-based relevance model that separates relevant and irrelevant data items. The model can be explored for visual model understanding. Moreover, the model visually conveys areas of uncertainty. The user is then able to refine the relevance feedback in areas of uncertainty, namely the decision boundary. Since our approach is focused on the creation of a relevance model reflecting the user's notion of relevance and thus in essence, not for exploratory analysis, we limit our approach to the representation of a user's fixed notion of relevance.

\smallskip
\noindent\textbf{Classifier Training:} A regular SOM is likely to create cells in which relevant and irrelevant items are mixed. We resolve this by proposing a hierarchical SOMs that allows for the expression of fine-grained differences in the user's notion of relevance. For this reason, we merge the concept of a tree with the concept of child SOMs presented by Sacha et al.~\cite{Sacha2018}, where a new SOM is calculated only with a subset of the dataset determined by the cell selection of a parent SOM. However, our algorithm creates a classifier automatically without any user interaction other than supplied labels. We automatically calculate a child SOM only from the data items assigned to the given cell $c$ if this cell exhibits a mixture of relevant of irrelevant greater then a threshold $m_t$, i.e., $MixRatio(c) > m_t$ with   
\begin{equation} 
  MixRatio(c) = min(|\mathpzc{L}^+_c| / |\mathpzc{L}_c|, |\mathpzc{L}^-_c| / |\mathpzc{L}_c|)
\end{equation}
The cell needs to contain enough data items in order for a child SOM to be useful. We model this circumstance by another threshold value $c_t$, such that the number of items in a cell $|E_c|$ must exceed $c_t$. Thus $c_t$ determines the split criterion. In \tool{}, the creation of SOM models is based on the supplied similarity measure, as determined by the \emph{Similarity Advisor}, and the relevance labels. The resulting SOM-based model can exhibit a tree structure (\autoref{fig:som-classifier} (1)). We limit the layout to $3\times3$ to leverage the projection of a SOM into 2D but not handle an excessive amount of children for a given parent in the classification tree.

\smallskip
\noindent\textbf{Classification of Data Items}
A classification of a given data item is performed recursively, similar to a decision tree. (1) Find the most similar neuron in the root SOM; (2) If the node has a child, perform the same action recursively on the child SOM; (3) If the SOM node has no child, classify the item as the predominant label of the respective cell, i.e., relevant or irrelevant; (4) If no label information is available for this node, use the next most similar cell with label information in that specific SOM.

\subsection{SOM Exploration and Refinement}

Our SOM-based visual classifier is visually explorable. It conveys its relevance decisions through multiple visual and interactive techniques. The main navigation happens in the visual classifier tree (\autoref{fig:som-classifier} (1)). Each SOM can be selected to examine it in detail. The currently active SOM is marked with a purple border. A purple dot highlights the parent SOM of the selected child SOM. The color coding of the grid in each SOM is intuitive, green signals a predominance of relevant items, red a predominance of irrelevant items. Yellow signals a mixture of relevant and irrelevant items, according to the $MixRatio$ of a cell. Such cells are likely to be recursively refined, as described in the previous section. We deliberately chose this encoding since it intuitively signals the relevance of data items from green over yellow to red gradient. \autoref{fig:som-classifier} (3) shows the classification outcome for data items assigned to a child SOM or individual cell. This allows us to detect whether a cell is on decision boundary.

To provide insight into the data items assigned to each node, we provide a range of stackable cell visualizations that can be selected in a user-defined order.

\smallskip
\lettrine[findent=.5em,nindent=0pt,lines=5]{\protect\includegraphics[height=17mm]{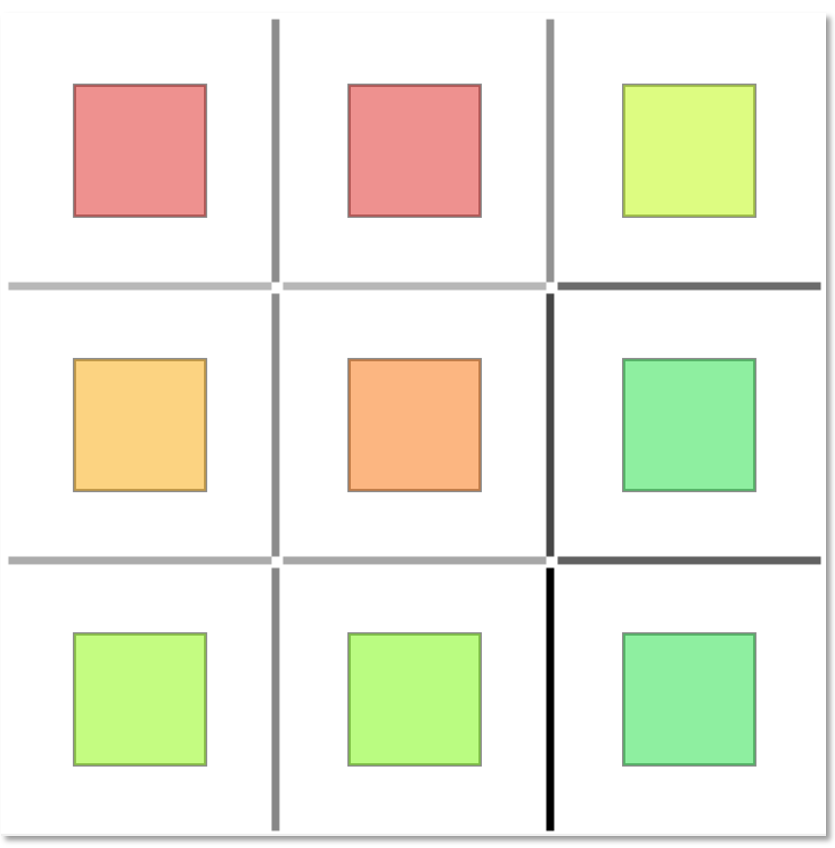}}{}
\textbf{Relevance Label Quality:} 
The label quality is depicted as colored squares on top of each node. We use the $MixRatio$ to determine the color and create a gradient from red over yellow to green; red is representing only irrelevant, green only relevant items within the cell and yellow implies an uncertain cell, i.e., decision boundary. A white dot signalizes that the cell contains not enough labeled data items, visually prompting users for more labels.

\smallskip
\lettrine[findent=.5em,nindent=0pt,lines=5]{\protect\includegraphics[height=17mm]{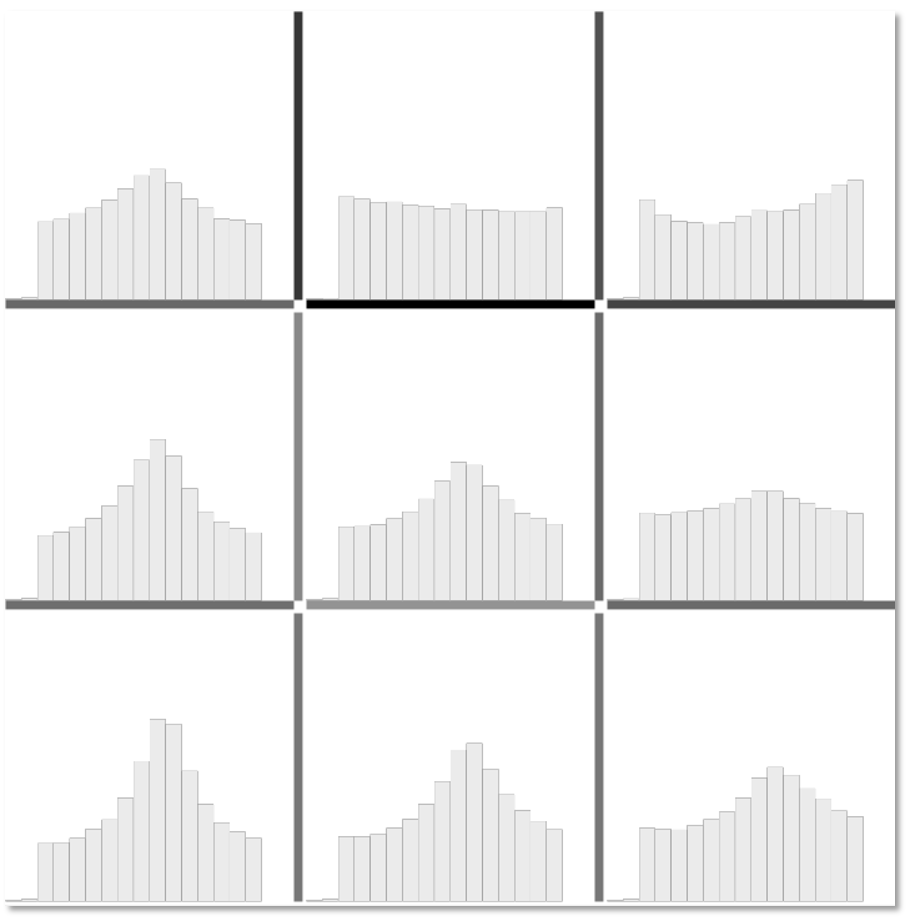}}{}
\textbf{Feature Histogram:} This layer displays the trained vector of the node. It can be used to judge the differences of SOM cells according to the currently recommended feature description. If the currently active FD is interpretable, like an FD derived from a color histogram, describing the color spectrum of an image, it can also hint at the properties of the contained data items. 

\smallskip

The user can also utilize two other layers, the quantization error (QE) \cite{Poelzlbauer2014} and the U-Matrix \cite{Ultsch1999}, to explore clusters of nodes that should be treated similarly by the model. Also, we support the user with detailed information about the number of assigned data items, relevant, irrelevant, and neutral data items. This information allows the user to judge the importance of a given node and the amount of information available to the model.

\subsection{Visual Active Learning with SOMs}

Cells with a low amount of labeled data items are uncertain. We measure this uncertainty with the $LabelRatio$ of a cell $c$. $|E_c|$ defines the number of items in a cell. Thus, we define the $LabelRatio$ as
\begin{equation} 
  LabelRatio(c) = (|\mathpzc{L}^+_c| + |\mathpzc{L}^-_c|) / |\mathpzc{E}_c|
\end{equation}
The model marks cells that do not have a child SOM with a white dot if $LabelRatio(c) < q_t$, where $q_t$ defines a threshold. A white dot signals uncertain neurons with a low label count to prompt the user to supply additional labels in these uncertain data regions. If the user does not supply an additional label by the suggested SOM node, the query formulated by an active learning system is generated from those marked nodes. For every node, the user can request details-on-demand in the form of a model-refinement dialog, similar to the annotation view, presented in~\autoref{sec:annotation-view}.

\section{Evaluation}\label{sec:evaluation}

\begin{figure*}[th]
    \centering
    \includegraphics[width=\linewidth]{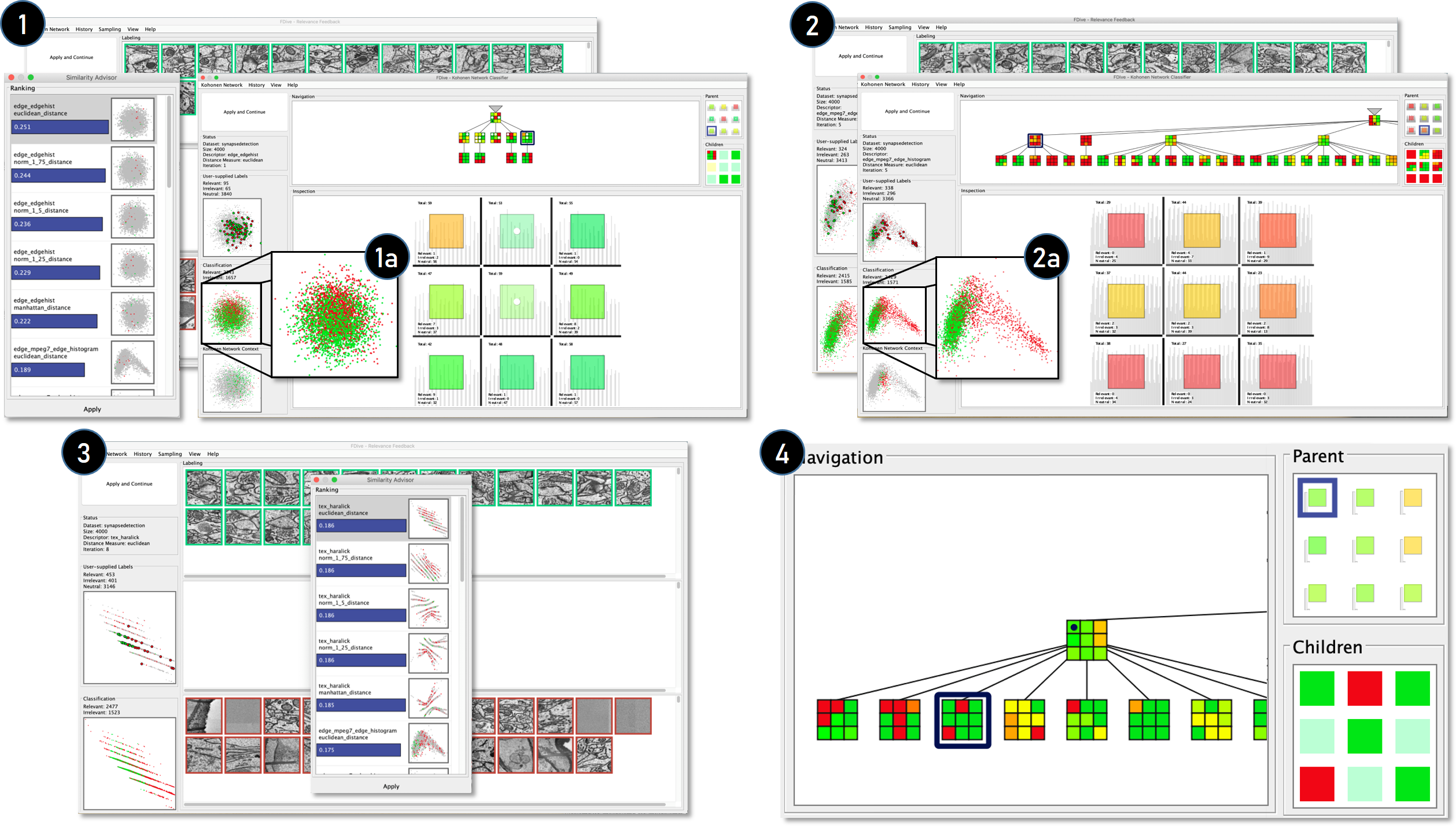}
    \caption{\tool{} learns to differentiate Electron Microscopy (EM) images containing synapses from images that do not. The domain expert found the classification model to be satisfactory after nine iterations. We show four key events in the model learning process. (1) The initial model is classifying the data very poorly, as presented by the scatter plot (1a) being very noisy and mixed. (2) The scatter plot shows a cleaner decision boundary (2a) and the model gets more complex, while the expert labels requested data items. (3) In the seventh iteration, the domain expert noticed that the \textsc{Haralick}~\cite{Haralick1973} FD combined with the Euclidean distance is recommended for the third time in a row, hinting at convergence for the similarity measure. (4) The last two iterations were spent exploring the model, observing and refining decision boundaries.}
    \label{fig:case-study}
\end{figure*}

Approaches involving relevance feedback are not straightforward to evaluate, as the results depend on both hidden and explicit user preferences and the definition of the learning components~\cite{eval-paper}. Therefore, we show its usefulness by applying it to a real-world use-case. We evaluate the general workflow, including the \emph{Similarity Advisor}, through a comparison to multiple feature selection techniques.

\subsection{Case Study: Synapse Detection}\label{subsec:case-study}

The goal of connectomics is to reconstruct the neural wiring diagram from Electron Microscopic (EM) images of the animal brain to improve the understanding of neuropathology and intelligence. A synapse is a functional structure that enables signal transfer from one neuron to the other, which connects individual neurons into a complex network. Manual labeling of synapses can be extremely hard because (1) there are approximately one billion synapses in a 1$mm^3$ cube of a mouse brain, and (2) the labeling of synapses requires expertise and cannot be crowdsourced. Therefore, a good labeling system of synapses should be semi-automatic and only provide informative samples to the domain experts to improve the labeling efficiency.  To showcase the effectiveness of our proposed approach, we applied the annotation system to a high-resolution EM image dataset generated by a multi-beam scanning electron microscope\footnote{Appendix A provides a visual overview of the Synapse Detection dataset.}. In total, there are 4,000 image patches, half of them containing a synapse at the center of the image, while the other half do not contain synapses. In this study, we show how our system helps experts classify synapse images and non-synapse images without any labeled training set and pre-specified domain knowledge.

CNN-based approaches have achieved state-of-the-art performance on image classification tasks~\cite{cnn,he2016deep}. However, there are still two main shortcomings of CNN-based methods. First, because the model space of CNNs can be huge, the model can easily overfit the training set and have poor performance on the test set, which requires a large training set. Second, the features extracted over convolutional layers are hard to interpret, which restricts the understanding of the discriminative features, especially for scientific applications where the expert wants to have a full understanding of the model.

Thus, we perform a case study\footnote{Appendix C shows all intermediate steps in HD images.} involving the classification of Electron Microscopy (EM) images of brain cells. A domain expert is tasked with the creation of a relevance model able to distinguish images depicting neuronal synapses. The domain expert has experience in the area of connectomics and the interpretation of EM images, including the identification of cell structures such as cell organelles and neuronal synapses. The study was conducted as a semi-structured interview. The case-study was performed after a training period. The expert performed a total of nine iterations to teach our relevance model the difference between EM images containing synapses and those which do not.
\autoref{fig:case-study} shows four key events in the model learning process.
After the initial annotation of 40 data items, the system suggested the \textsc{Edgehist} FD. The expert finished the first iteration by labeling data items in cells with a white dot. A total of 95 images were annotated as relevant and 65 as irrelevant. In the second iteration, the system suggested the \textsc{Tamura} FD. The expert labeled 63 images as relevant and 57 as irrelevant. In the third iteration, the system suggested the \textsc{Tamura} FD again. In the fourth and fifth iteration, the \textsc{MPEG7 Edge Histogram} FD was suggested. In iteration six to nine the system consistently suggested the \textsc{Haralick} FD point at convergence on this specific FD. The expert followed the recommendations of the \emph{Similarity Advisor} in every iteration, finishing after the ninth iteration.

In the first three iterations, the system indicated uncertain cells. In later iterations, we are able to check the distribution of samples in a SOM on the scatter plots to see if they are still mixed up. In the end, it notified the expert that it has enough labels, such that no further inspection or labeling is necessary. After several iterations of labeling, the expert noticed that samples are separated in the classification scatter plot, and, when inspecting the individual nodes pertaining to a data region, the labels of similar data items were matching. From the root node to the leaf nodes, he was able to see a trend towards purity. Therefore, when the uncertainty indicators (i.e., white dot) disappears, the nodes with mixed colors are more appealing to be labeled. The inspection of nodes was helpful to the expert to validate whether a set of samples spread out on the scatter plots and thus do not form a coherent cluster. When inspecting a cell colored in yellow, the expert was able to see decision boundaries. Subsequently, the expert labeled ten queried samples to refine the decision boundary. After labeling one node, the color of the node itself and its sibling nodes may change, and the expert was able to verify the impact. The expert noted that the appearance of the scatter plot changed several times at the initial iteration and that the relevant and irrelevant samples on the scatter plots were mixed and not forming a coherent cluster. However, after several iterations, the model converges to a specific similarity measure, and samples become more separable on the scatter plots.

With \tool{} we can learn to distinguish and extract relevant patterns from a large high dimensional dataset, in this case, EM images depicting synapses, using a sparse amount of labels. Whenever a new label is applied, the system conveys its impact visually. The relevance model is visually explorable and refinable such that the expert was able to asses the model quality and the convergence towards a useful relevance model.

\subsection{Quantitative Framework Evaluation}\label{subsec:quant-evaluation}

\begin{table}[tb]
\setlength{\tabcolsep}{5pt}
\centering
{\small
\begin{tabular}[c]{cccccccc}
\toprule
\multicolumn{2}{c}{\textbf{Labeling}} & \multicolumn{6}{c}{\textbf{Results} (larger is better)} \\
\midrule
\multirow{3}{*}{\shortstack[c]{\textbf{Target} \\ Example} }& \textbf{\#Labels} & \multicolumn{3}{c}{\textbf{Best Selected FD}} & \multicolumn{3}{c}{\textbf{Best Ranked Original FD}}  \\
& $\mathpzc{L}^+$ / $\mathpzc{L}^-$ & \multicolumn{3}{c}{Baseline} & \multicolumn{3}{c}{\tool{}} \\
& each & $k$ = 1 & $k$ = 3 & $k$ = 5 & $k$ = 1 & $k$ = 3 & $k$ = 5 \\
\toprule
\multirow{5}{*}{\shortstack[c]{\\ \includegraphics[width=.12\linewidth]{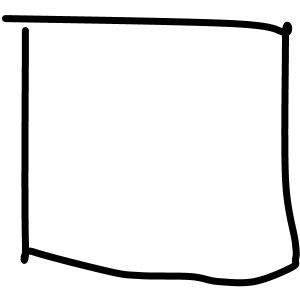}}}
& \tecw{25} & \tecb{.359} & \tecb{.397} & \tecb{.410} & \tec{.268} & \tec{.317} & \tec{.312} \\
& \tecw{50} & \tecb{.398} & \tecb{.464} & \tecb{.449} & \tec{.238} & \tec{.330} & \tec{.330} \\
& \tecw{75} & \tecb{.326} & \tecb{.436} & \tecb{.490} & \tec{.215} & \tec{.295} & \tec{.328} \\
& \tecw{100} & \tecb{.350} & \tecb{.407} & \tecb{.465} & \tec{.239} & \tec{.321} & \tec{.347} \\
& \tecw{125} & \tecb{.437} & \tecb{.516} & \tecb{.494} & \tec{.250} & \tec{.328} & \tec{.368} \\
\hline
\vspace{-.22cm}\\
\multirow{5}{*}{\shortstack[c]{\\ \includegraphics[width=.12\linewidth]{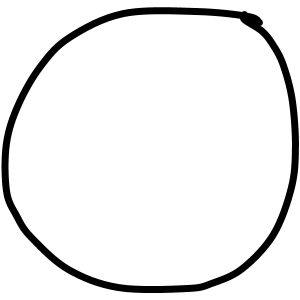}}}
& \tecw{25} & \tecb{.363} & \tecb{.368} & \tecb{.320} & \tec{.272} & \tec{.264} & \tec{.239} \\
& \tecw{50} & \tecb{.399} & \tecb{.444} & \tecb{.426} & \tec{.296} & \tec{.292} & \tec{.279} \\
& \tecw{75} & \tecb{.461} & \tecb{.533} & \tecb{.542} & \tec{.286} & \tec{.309} & \tec{.292} \\
& \tecw{100} & \tecb{.539} & \tecb{.611} & \tecb{.567} & \tec{.306} & \tec{.338} & \tec{.323} \\
& \tecw{125} & \tecb{.507} & \tecb{.600} & \tecb{.602} & \tec{.304} & \tec{.357} & \tec{.345} \\
\hline
\vspace{-.22cm}\\
\multirow{5}{*}{\shortstack[c]{\\ \includegraphics[width=.12\linewidth]{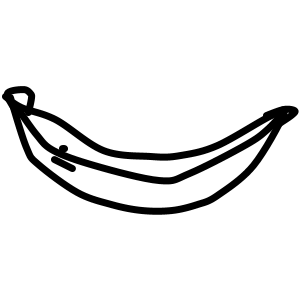}}}
& \tecw{25} & \tec{.212} & \tec{.222} & \tec{.224} & \tecb{.556} & \tecb{.566} & \tecb{.490} \\
& \tecw{50} & \tec{.303} & \tec{.310} & \tec{.306} & \tecb{.561} & \tecb{.574} & \tecb{.578} \\
& \tecw{75} & \tec{.323} & \tec{.351} & \tec{.362} & \tecb{.605} & \tecb{.619} & \tecb{.626} \\
& \tecw{100} & \tec{.473} & \tec{.526} & \tec{.507} & \tecb{.529} & \tecb{.595} & \tecb{.586} \\
& \tecw{125} & \tec{.363} & \tec{.447} & \tec{.469} & \tecb{.522} & \tecb{.585} & \tecb{.606} \\
\hline
\vspace{-.22cm}\\
\multirow{5}{*}{\shortstack[c]{\\ \includegraphics[width=.12\linewidth]{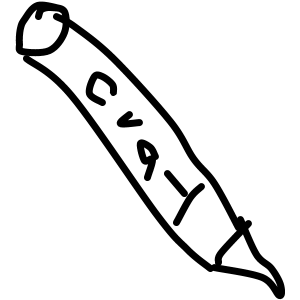}}}
& \tecw{25} & \tec{.152} & \tec{.170} & \tecb{.187} & \tecb{.166} & \tecb{.174} & \tec{.153} \\
& \tecw{50} & \tec{.175} & \tec{.157} & \tec{.171} & \tecb{.192} & \tecb{.216} & \tecb{.222} \\
& \tecw{75} & \tec{.180} & \tec{.192} & \tec{.184} & \tecb{.197} & \tecb{.205} & \tecb{.202} \\
& \tecw{100} & \tec{.160} & \tec{.179} & \tec{.186} & \tecb{.192} & \tecb{.203} & \tecb{.194} \\
& \tecw{125} & \tecb{.173} & \tecb{.186} & \tecb{.181} & \tec{.135} & \tec{.142} & \tec{.145} \\
\hline
\vspace{-.22cm}\\
\multirow{5}{*}{\shortstack[c]{\\ \includegraphics[width=.12\linewidth]{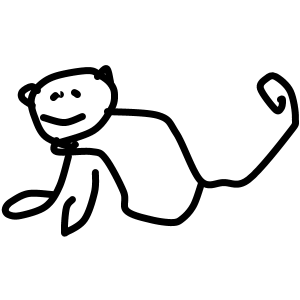}}}
& \tecw{25} & \tecb{.179} & \tecb{.169} & \tecb{.173} & \tec{.096} & \tec{.105} & \tec{.101} \\
& \tecw{50} & \tec{.162} & \tec{.165} & \tec{.176} & \tecb{.183} & \tecb{.247} & \tecb{.253} \\
& \tecw{75} & \tecb{.197} & \tec{.201} & \tec{.215} & \tec{.180} & \tecb{.222} & \tecb{.254} \\
& \tecw{100} & \tec{.180} & \tec{.176} & \tec{.191} & \tecb{.186} & \tecb{.245} & \tecb{.273} \\
& \tecw{125} & \tecb{.193} & \tec{.209} & \tec{.210} & \tec{.180} & \tecb{.235} & \tecb{.262} \\
\bottomrule
\\
\end{tabular}
}
\caption{We compare the $F_1$ scores for different k-NN classifiers. Our heuristic approach performs better for analysis targets with a higher complexity (i.e. \emph{banana}, \emph{crayon} and \emph{monkey}) than state-of-art feature selection algorithms that can draw features from all available feature descriptors (4694 features). It performs worse for less complex patterns (i.e. \emph{square} and \emph{circle}).}
\label{tab:quant-evaluation}
\vspace*{-1.5em}
\end{table}

This evaluation compares the best best-breed-competitor generated by 3 algorithms and 4 different FD sizes against our ``one-shot'' \emph{Similarity Advisor} result. Comparing a recombination of all features with the \emph{Similarity Advisor} using only the predefined feature descriptors make this evaluation biased against our approach. However, we were still able to outperform the best best-breed-competitor {in 36 out of 75 cases}. We evaluate \tool{} on the following options and parameter settings with the central goal to show the usefulness of ranking pattern-based similarity measures for model learning. We provide a comprehensive overview of the results in \autoref{tab:quant-evaluation}\footnote{Appendix B contains complete records of all experiments.}. The basis for all experiments is the \emph{Quick, Draw!} dataset~\cite{QuickDraw2016}\footnote{Appendix A provides a visual overview of the \emph{Quick, Draw!} dataset~\cite{QuickDraw2016}.}. We reduced the dataset to 4500 images consisting of 150 sketches for each of the 30 labels, describing the depicted objects. We choose the labels \emph{square}, \emph{circle}, \emph{banana}, \emph{crayon}, and \emph{monkey}. These labels cover a variety of shapes with different complexity. We assume each label as a specific analysis target. For each of the target labels, we label progressively more items as relevant and irrelevant. The progression is 25/25, 50/50, 75/75, 100/100, and 125/125 for $\mathpzc{L}^+$ / $\mathpzc{L}^-$. This sequence represents an increase in the available labels through the iteration cycle. To verify the validity of the similarity measure ranking, we train a k-NN classification model. We chose k-NN, because it is fully automatic and represents an intuitive classification model. We select three parameters for $k$, namely 1, 3, and 5. To make our results invariant to the feature selection technique, we conducted our experiments using the ReliefF algorithm, a Linear Ranking Ensemble consisting of ten Recursive Elimination SVMs, and a regular Recursive Elimination SVM. These techniques are described in \autoref{subsec:feature-selection}. Those algorithms rank features according to their significance. We choose subsets of different lengths, namely 5, 10, 15, and 20. We perform a feature selection on the concatenation of all FDs (4694 features), resulting in recombination of different features, according to the significance assigned by the feature selection algorithm. This approach creates 12 (= 3 algorithms $\times$ 4 sizes) recombined FDs for each label and label count (i.e., table row). We determine the $F_1$ score of the trained k-NN for each $k$ with all recombined FDs and all distance functions, yielding 60 (= 12 selected FD $\times$ 5 similarity coefficients) $F_1$ scores for each $k$ parameter of the k-NN classifier. \autoref{tab:quant-evaluation} shows the best score out of 60 for a given $k$ in the three columns titled ``Best Selected FD'', serving as the benchmark. We compare this score to the single one resulting from a classification based on the best-ranked similarity measure according to the \emph{Similarity Advisor}. All FDs are in their original state and combined with all available distance functions. The \emph{Similarity Advisor} ranks the similarity measures based on the same label information as available to the feature selection. \autoref{tab:quant-evaluation} shows the $F_1$ score for a given $k$ for the best ranked similarity measure in the three rightmost columns titled ``Best Ranked Original FD''.

\medskip

Generally, we found that our the suggested similarity measure performs on a similar level than the best feature selection created by the feature selection algorithms. It outperforms the feature selected FD in all scenarios involving the \emph{banana} label and in 11 out of 15 scenarios pertaining to the \emph{crayon} label. The best-ranked \emph{Similarity Measure} is outperformed in scenarios where the analysis target is a less complex shape (i.e., \emph{square} and \emph{circle}. In case of the \emph{monkey} label, our ranked FD can achieve similar result than the selected FD with 50 or more labeled instance for each of $\mathpzc{L}^+$ / $\mathpzc{L}^-$. Given that we compare 60 feature selection-based similarity measures to our single best ranked fixed-FD similarity measure, we can say that the similarity advisor is an efficient and effective method for the evaluation of similarity measures and that the best-ranked measure helps in the creation of a relevance model.

\section{Discussion and Future Work}\label{sec:discussion}

With \tool{}, we provide a technique which allows for the iterative learning of a relevance model, including the definition of a useful similarity measure. In the case of \tool{}, a similarity measure comprised of a feature descriptor and a distance function. The visual guidance of the SOM-based relevance model to uncertain classification near decision boundaries improved the understanding and quality of the model. We show that the continuous evaluation of the similarity measure benefits the iterative creation of relevance models, helping them to converge towards increasingly useful results.

\medskip

One area of improvement noted by the expert was that, upon change of the similarity measure, the relevance model changes its layout, requiring the analyst to relearn it. For this reason, the mapping of different model representations into various feature spaces would allow us to explore the impact of a changed feature space on the model. Making this effect accessible would further the understanding of the feature space and underlying data distribution.

\medskip

We plan to extend the general concept of the \emph{Similarity Advisor} to other types of distance functions, removing the limitation to vector spaces implied by the $L^p$ Minkowski family of distance measures. This extension would allow us to use other distance functions, such as Cosine, Canberra, or Clark distance. Analysts apply these measures often in specific scenarios and domains. The automatic detection of a distance function would replace the need for an expert, removing the bias introduced through the single fixed distance function. Additionally, we want to explore the application of the \emph{Similarity Advisor} in different contexts, such as the validation of feature weightings or the design of feature descriptors based on prototypical representations of the described properties. In this instance, the \emph{Similarity Advisor} could serve as a \emph{concept validator}. Feature descriptors can be linked to visualization types. Through a technique similar to the \emph{Similarity Advisor}, it should be possible to suggest other data representations, such as switching from a scatter plot representation to a parallel coordinate plot. An automatic suggestion of a useful visualization would add another step to a generalized analysis workflow, where many choices an analyst or even system designer can make is automatically assessed and supported. We layout the SOM-based relevance model in a tree structure, because it is explainable and an intuitive way of reading a classifier. Techniques introduced by Sacha et al.~\cite{Sacha2018} can be used to enhance its descriptive ability. This addition can lead to novel SOM interactions focused on classification rather than exploratory cluster analysis.

\medskip

We discuss scalability on two levels. First, we discuss the computational effort of \emph{Similarity Advisor}. The main computational effort lies in the required preprocessing to transform the feature spaces and distance functions into a comparable format. The transformations are parallelizable. The complexity is determined by the dataset size. The complexity of the \emph{Inter-Group-Distance} calculation is determined by the number of supplied labels. However, this relationship is linear. 
Second, we discuss the scalability limit of the complete FDive approach.
The main limit approach is the creation of the SOM-based relevance model. However, the results of a previous iteration cycle can be reused in the subsequent cycles. One issue that we found was that the tree representation of the SOM-based model can become very wide. Here we have to consider a tradeoff between the size of the SOM and the associated data partitioning properties and the number of child SOMs leading to a broad tree. We found that a $3 \times 3$ SOM is an acceptable size for the SOMs since it is a size where the 2D projection property has a notable effect.

\section{Conclusion}\label{sec:conclusion}

The extraction of interesting patterns from large high-dimensional datasets is a challenging task. With \tool{}, we present a workflow for the creation of relevance models based on \emph{pattern-based similarity measures}. The system ranks similarity measures according to how well they separate relevant from irrelevant data. Our SOM-based relevance model is interactively explorable and guides the user to uncertain areas, i.e., decision boundaries. We evaluated our technique with a real-world case study in which we show that \tool{} can reflect the complex differences between electron microscopy images showing synapses of neurons or other brain cell structures. Our comparison to feature selection shows that \tool{}'s \emph{Similarity Advisor} serves as a useful metric to evaluate the discriminative ability of feature descriptor and distance function combinations. With \tool{}, we introduce the concept of continuous \emph{Similarity Advisor} assessment during the learning process of a relevance model. The \emph{Similarity Advisor} concept is applicable areas where the user expresses his relevance for specific data items and can improve the results of the given task. The full \tool{} approach allows the creation of relevance models for a complex task while providing the user with valuable insights about the learning process, such as the underlying similarity measure and the model properties, including the judgment of classification results in areas of high uncertainty.

\acknowledgments{This work was funded by the Deutsche Forschungsgemeinschaft (DFG, German Research Foundation) within the projects A03 of TRR 161 (Project-ID 251654672) and Knowledge Generation in VA (Project-ID 350399414). We thank Michael Blumenschein and the anonymous reviewers for their valuable feedback.}

\bibliographystyle{abbrv-doi-hyperref-narrow}

\bibliography{ms}
\end{document}